%% file: ijcai_2024.tex
\documentclass{article}
\usepackage{ijcai_2024}

\usepackage{times}
\usepackage{soul}
\usepackage{url}
\usepackage[hidelinks]{hyperref}
\usepackage[utf8]{inputenc}
\usepackage[small]{caption}
\usepackage{graphicx}
\usepackage{amsmath}
\usepackage{amsthm}
\usepackage{booktabs}
\usepackage{algorithm}
\usepackage{algorithmic}
\usepackage[switch]{lineno}
\usepackage{booktabs}
\usepackage{makecell}
\usepackage{pifont}
\input{setting}

\usepackage{url}

\title{\textsc{Guide}: A Guideline-Guided Dataset for Instructional Video Comprehension}

\author{
Jiafeng Liang$^1$
\and
Shixin Jiang$^1$
\and
Zekun Wang$^1$
\and
Haojie Pan$^3$
\and
Zerui Chen$^1$
\and
Zheng Chu$^1$
\and
\\Ming Liu$^{1,2}$\thanks{~~Corresponding Author.}
\and
Ruiji Fu$^3$
\and
Zhongyuan Wang$^3$
\And
Bing Qin$^{1,2}$
\affiliations
$^1$Harbin Institute of Technology, Harbin, China\\
$^2$Peng Cheng Laboratory, Shenzhen, China\\
$^3$Kuaishou Technology, Beijing, China\\
\emails
\{jfliang, sxjiang, zkwang, zrchen, zchu, mliu, qinb\}@ir.hit.edu.cn
}

\begin{document}
\maketitle

\input{Sections/0_abstract}
\input{Sections/1_introduction}

\input{Sections/2_related_work}
\input{Sections/3_dataset}
\input{Sections/4_experiments}
\input{Sections/5_conclusion}

\input{Sections/6_ethics}

\section*{Acknowledgements}
We thank anonymous reviewers for their insightful feedback that helped improve the paper.
The research in this article is supported by the National Key Research and Development Project (2021YFF0901602), the National Science Foundation of China (U22B2059, 62276083), and Shenzhen Foundational Research Funding (JCYJ20200109113441941), Major Key Project of PCL (PCL2021A06).

\bibliographystyle{named}
\bibliography{ijcai_2024}

\clearpage
\appendix
\input{Sections/7_appendix}

\end{document}

%% file: setting.tex
\usepackage{times}
\usepackage{latexsym}
\usepackage{graphicx}
\usepackage{arydshln}
\usepackage{subfigure}
\usepackage{makecell}
\usepackage{colortbl}
\usepackage{pifont}
\usepackage{amssymb}

\usepackage[T1]{fontenc}
\usepackage{xcolor}
\definecolor{mygoldenrod}{RGB}{184,134,11}
\definecolor{myyellow}{RGB}{230,213,37}
\definecolor{myyellow2}{RGB}{251,239,232}
\definecolor{myblue}{RGB}{78,186,212}
\definecolor{myblue2}{RGB}{153,170,237}
\usepackage[utf8]{inputenc}

\usepackage{microtype}

\usepackage{array}
\usepackage{pifont}
\usepackage{tabularx}
\usepackage{adjustbox}
\usepackage{multirow}
\usepackage{enumitem}
\usepackage{xspace}
\usepackage{tcolorbox}
\usepackage{xparse}
\usepackage{soul}
\usepackage{booktabs,amsfonts,dcolumn}
\usepackage{url}
\usepackage{amsmath,amsthm,amsfonts,amssymb,bm,stmaryrd}
\usepackage{breqn}

%% file: Sections/0_abstract.tex
\begin{abstract}

There are substantial instructional videos on the Internet, which provide us tutorials for completing various tasks.
Existing instructional video datasets only focus on specific steps at the video level, lacking experiential guidelines at the task level, which can lead to beginners struggling to learn new tasks due to the lack of relevant experience.
Moreover, the specific steps without guidelines are trivial and unsystematic, making it difficult to provide a clear tutorial.
To address these problems, we present the \textsc{\textbf{Guide}} (\textbf{G}uideline-G\textbf{uide}d) dataset\footnote{\url{https://guide-ijcai2024.github.io/}}, which contains 3.5K videos of 560 instructional tasks in 8 domains related to our daily life.
Specifically, we annotate each instructional task with a guideline, representing a common pattern shared by all task-related videos.
On this basis, we annotate systematic specific steps, including their associated guideline steps, specific step descriptions and timestamps. 
Our proposed benchmark consists of three sub-tasks to evaluate comprehension ability of models:
(1) Step Captioning: models have to generate captions for specific steps from videos.
(2) Guideline Summarization: models have to mine the common pattern in task-related videos and summarize a guideline from them. 
(3) Guideline-Guided Captioning: models have to generate captions for specific steps under the guide of guideline. 
We evaluate plenty of foundation models with \textsc{Guide} and perform in-depth analysis.
Given the diversity and practicality of \textsc{Guide}, we believe that it can be used as a better benchmark for instructional video comprehension.

\end{abstract}

%% file: Sections/1_introduction.tex
\section{Introduction}
 \begin{figure}[t]
    \centering
    \includegraphics[width=\linewidth]{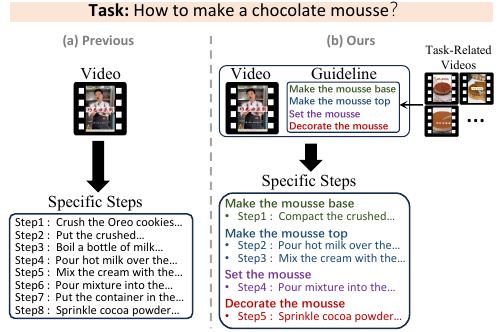}
    \caption{The steps in the previous dataset were very trivial and unsystematic, making it difficult for beginners to learn. In contrast, our dataset provides structured guideline-guided steps. Such guideline is a common pattern shared by all videos related to the same task.}
    \label{fig:compare}
\end{figure}

Instructional videos guide learners how to accomplish multi-step tasks such as cooking, making up and embroidering, repairing, or creating new objects.
Recently, numerous instructional video datasets have been proposed~\cite{hirest,coin,youcook2,crosstask}.
As shown in Figure~\ref{fig:compare} (a), these datasets solely focus on fine-grained annotations, leading to trivial and unsystematic step captions, making it difficult to provide clear tutorial guidance.
Moreover, while many instructional videos pertain to the same task, there are significant differences in the details and sequence of their steps, which increases the difficulty for beginners to learn.
If there exists a model capable of analyzing various videos of the same task and organizing steps into a hierarchical structured tutorial, it will accelerate learning progression for beginners.

\begin{figure*}[t]
    \centering
    \includegraphics[width=\linewidth]{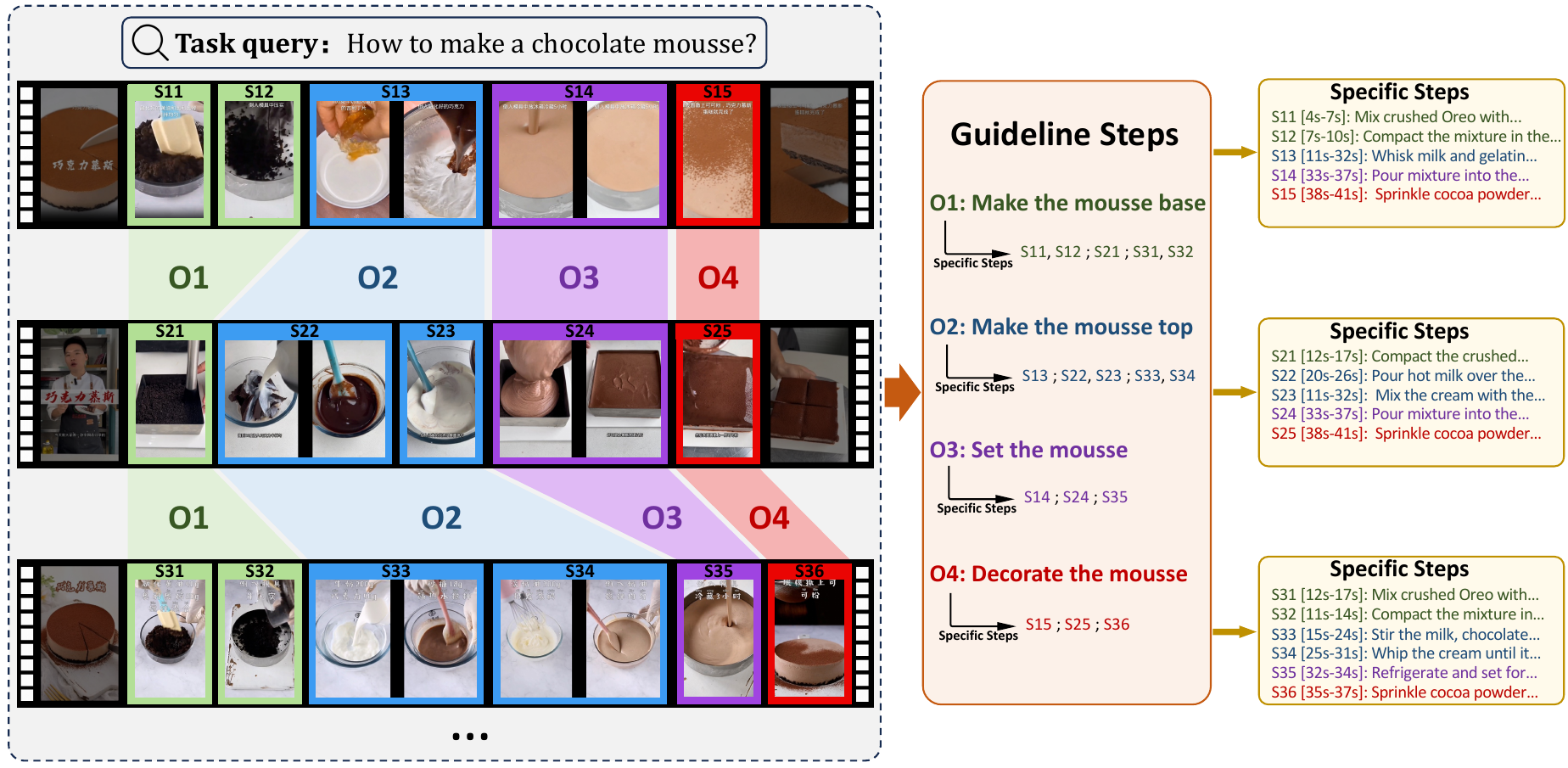}
        \caption{Overview of the \textsc{Guide} dataset. The \textsc{Guide} consists of 560 task queries, each containing an average of 6.2 task-related videos. These instructional videos are divided into specific steps with timestamps and text descriptions (yellow area). Additionally, each task contains a set of guideline steps representing a common pattern shared by all task-related videos (purple area).}
    \label{fig:dataset}
\end{figure*}

Our inspiration comes from two aspects.
First, according to educational psychology~\cite{xinlixue}, learners are always confused when learning an unfamiliar and challenging task because they lack relevant experience.
Although previous datasets provide specific steps, their complexity would exceed learners' cognitive load.
Thus, a clear guideline can help learners understand the task more efficiently.
Second, planning a procedure from an instructional video is to complete a specific guideline ({\it i.e.}, a procedure matches a clear intention).
Since a guideline usually involves specific steps, they can be used to support the procedure generating (shown in Figure~\ref{fig:compare} (b)).

To support these, we introduce \textsc{\textbf{Guide}}, a \textbf{G}uideline-G\textbf{uide}d dataset for instructional video comprehension.
We propose a three-stage dataset construction pipeline on instructional video from video platform\footnote{Chinese short video platform: Kuaishou.}, collecting high-quality annotations.
The \textsc{Guide} contains three annotations (shown in Figure~\ref{fig:dataset}):
(1) 560 queries: each query represents an instructional task and contains an average of 6.2 task-related videos (total of 3.5K videos),
(2) 15K step segments: each video is divided into an average of 4.3 specific step segments with corresponding timestamps and text description, and,
(3) 560 guidelines: each instructional task contains a set of guideline steps that represent a common pattern of the task.
Moreover, each specific step has its corresponding guideline step.

In \textsc{Guide}, we propose three challenging sub-tasks for instructional video analysis.
(1) Step Captioning: models have to generate captions for specific steps from videos.
(2) Guideline Summarization: models have to mine the common pattern in task-related videos and summarize a guideline from them.
(3) Guideline-Guided Captioning: models have to generate captions for specific steps under the guide of guidelines.
We benchmark various video foundation models and language foundation models (utilize video transcription instead of visual information), including VideoChat~\cite{videochat}, Video-LLaMA~\cite{videollama}, mPLUG-Owl~\cite{mplug}, GPT-3.5-turbo~\cite{chatgpt}, GPT-4~\cite{gpt4}, Vicuna~\cite{vicuna} and Flan-T5~\cite{flant5}. 
We also evaluate the performance of humans on the \textsc{Guide} for a better comparison.

The experimental results demonstrate that both video and language foundation models are struggle in all three sub-tasks.
We initially observed that the specific steps generated under the ground-truth guideline guide are clearer and more accurate, indicating that the accurate guideline is helpful for generating instructional steps.
Then, we explore the source of the ability to mine the guideline from multiple videos with different training settings.
The results show that the single-video understanding ability is the basis of learning multiple videos, indicating that more pre-training and fine-tuning data is necessary.
Subsequently, we investigate the bottleneck of video foundation models. 
The model demonstrates significant performance degradation compared to their text-only counterparts, indicating that more specialized visual encoders and visual-language bridges are needed to represent temporal procedures better.
Finally, we perform a human evaluation demonstrating our dataset's promising applications in real-world scenarios.
To summarize our contributions:

\begin{itemize}

\item We introduce a novel guideline-guided instructional video comprehension dataset \textsc{Guide}, containing task-level guideline annotations and video-level systematic specific step annotations.

\item We design three challenging sub-tasks in \textsc{Guide} for instructional video analysis, namely step captioning, guideline summarization and guideline-guided captioning.

\item We benchmark various foundation models and conduct extensive analyses to provide detailed insights.

\end{itemize}

%% file: Sections/2_related_work.tex
\section{Related Work}
\begin{figure*}[t]
    \centering
    \includegraphics[width=\linewidth]{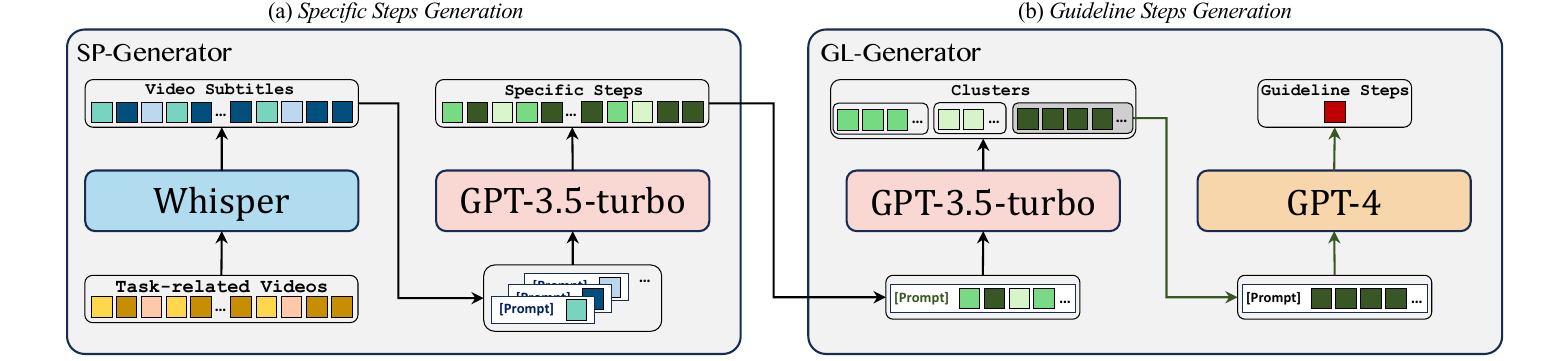}
    \caption{Overview of Automatic Annotation. (a) Transcribing the video into textual subtitles and generating specific steps based on subtitles. (b) Clustering the task-related videos and generating a set of guideline steps for the cluster with the highest number of videos.}
    \label{fig:gpt}
\end{figure*}

\subsection{Instructional Video Comprehension}
Understanding instructional videos presents a significant challenge, primarily due to their inherent procedural temporal structure.
Recently, many researchers have studied the analysis of instructional videos~\cite{stepformer,Vid2Seq,cap1}.
For instance, Yang {\it et al.}~\shortcite{Vid2Seq} propose an end-to-end framework that augments a language model to predict timestamps and descriptions of steps seamlessly.
Gu {\it et al.}~\shortcite{vc4} propose a two-stream transformer, which constructs a video scene graph~\cite{gtr} for video captioning by retrieving additional knowledge.
However, the steps generated by these methods are trivial and unsystematic due to the lack of guidelines, making it difficult for people to learn.
While some approaches~\cite{vcl1,vcl2} extract the guideline by analyzing correlations between instructional videos, they do not explore the help of the guideline for generating step captions.

\subsection{Instructional Video Datasets}
Existing instructional video datasets can be categorized into two types: action detection datasets~\cite{crosstask,coin,breakfast} and step caption datasets~\cite{youcook2,hirest,kitchen}.
The former are predominantly employed for video segmentation and action recognition tasks, while the latter are used for video segmentation and step captioning tasks.
For instance, COIN~\cite{coin} predefines many actions and assigns these actions to instructional videos to describe procedural processes.
HIREST~\cite{hirest} segments each video based on instructional steps and manually annotates captions for the steps.
However, these datasets primarily focus on fine-grained annotations, leading to trivial and unsystematic step captions, making it difficult to provide clear tutorial guidance.
Furthermore, many instructional videos related to the same tasks exhibit significant differences in specific procedures, increasing the difficulty for beginners to learn.
In this paper, Our \textsc{Guide} dataset provides guideline annotations, representing the common pattern across multiple task-related videos. 
In addition, we annotate guideline-guided specific steps to improve the systematic nature of the data, which reduces the learning difficulty.

%% file: Sections/3_dataset.tex
\section{Dataset}
\subsection{Overview}
\input{Tables/table_dataset_comparison}
In this section, we introduce our instructional video comprehension dataset, \textbf{\textsc{Guide}}.
Initially, we describe the three-stage dataset construction pipeline.
Subsequently, we provide an overview of the data statistics.
Lastly, we introduce the three novel sub-tasks we proposed to comprehensively evaluate foundation models based on our dataset.

\subsection{Dataset Construction Pipeline}
\textsc{Guide} dataset construction pipeline contains three stages: video collection, automatic annotation, and manual annotation.
In the Appendix~\ref{app:Dataset Construction Pipeline}, we provide more details for each stage.
\paragraph{Video Collection} 
In this stage, we aim to collect a large number of high-quality instructional videos. 
To ensure the widely-used of the \textsc{Guide}, we collect videos from 560 different instructional tasks across the 8 most common domains in our daily life.
We require annotators to collect videos containing explicit instructional steps and clearly defined time boundaries between these steps.
To further enhance the practicality of the dataset, we also require the collected videos that include detailed video subtitles, {\it{i.e.}}, each step accompanied by corresponding voice explanations. 
More details are provided in Appendix~\ref{app:Dataset Construction Pipeline}.

\paragraph{Automatic Annotation} 
As illustrated in Figure~\ref{fig:gpt}, the automatic annotation framework contains two stages: \textbf{Specific Steps Generation} and \textbf{Guideline Steps Generation}.

\textbf{In the specific steps generation}, we utilize the \textsc{SP-Generator} module, comprising Whisper~\cite{whisper} and GPT-3.5-turbo~\cite{chatgpt}, to automatically generate the specific steps for each video.
Given an instructional task query $Q$, which contains $n$ related videos $\left \{ v_{1},v_{2},...,v_{n}  \right \}$.
We first use Whisper to generate video subtitles as $A_{Q}=\left \{ a_{1},a_{2},...,a_{n}  \right \}$.
Subsequently, we feed these subtitles along with their occurrence time and our carefully crafted prompt to GPT-3.5-turbo, enabling it to generate specific steps $s$ and corresponding timestamps $t$ for each video.
Our \textsc{SP-Generator} can be formulated as:
 \begin{equation}
      SPsteps = \textsc{SP-Generator}(prompt,A_{Q})
 \label{eq:SP-Generator}
 \end{equation}
where $SPsteps = \left \{\left [ s,t \right ]_{1},\left [ s,t \right ]_{2},...,\left [ s,t \right ]_{n}   \right \}$. 
More details are provided in Appendix~\ref{app:Dataset Construction Pipeline}.
 
\textbf{In the guideline steps generation}, we aim to extract a set of guideline steps for each task.
Actually, extracting a shared guideline from all task-related videos is challenging due to the coarse granularity of task queries in the video database.
Specifically, despite many videos sharing the same query, they exhibit significant variations in specific content.
For instance, numerous videos fall under the task query `{\it{Making Crayfish}}', but variations in ingredients and procedures lead to diverse methods, such as `{\it{Spicy and Numbing Crayfish}}', `{\it{Fragrant and Spicy Crayfish}}' and `{\it{Garlic Crayfish}}'.
We try to involve annotators in clustering videos based on the video content during the video collection stage, with the objective of identifying a single cluster that best represents the task query.
However, manually clustering a large number of videos is challenging, and there are significant differences in subjective interpretations among annotators, making it difficult to establish clear clustering rules.
 
Hence, we utilize the \textsc{GL-Generator} module, comprising GPT-3.5-turbo~\cite{chatgpt} and GPT-4~\cite{gpt4}, to cluster the task-related videos and generate corresponding guideline steps automatically.
We feed $SPstep$ and crafted prompts into GPT-3.5-turbo to cluster the videos based on the content of the specific steps and their sequential order.
Then, we retain only the cluster $SPstep^{*}$ with the maximum number, which includes $m$ videos:
 \begin{equation}
      SPstep^{*} = \mathrm{Max} \left ( \mathrm{Cluster} \left ( SPstep \right )\right)
 \label{eq:summarizer}
 \end{equation}
where $SPstep^{*}$ = $\left \{ \left [ s,t \right ]_{1}, \left [ s,t \right ]_{2},...,\left [ s,t \right ]_{m} \right \}$.
Finally, we utilize GPT-4 to generate a set of guideline steps $GLstep$ for the current instructional task, as we find during the testing process that GPT-4 is capable of generating more comprehensive and common guideline steps compared to GPT-3.5-turbo. Our \textsc{GL-Generator} can be formulated as:
 \begin{equation}
  GLstep = \textsc{GL-Generator}(prompt,SPstep^{*})
 \label{eq:summarizer}
 \end{equation}
More details are provided in Appendix~\ref{app:Dataset Construction Pipeline}.
\paragraph{Manual Annotation}
The results of automatic annotation cannot be regarded as the final annotations.
GPT-3.5-turbo generates timestamps for each specific step based on the video subtitles' occurrence time.
However, these timestamps are inaccurate because the steps in the video and the voice explanation may not coincide, and due to the lack of information in the subtitles, the specific steps may not be complete.
In addition, despite providing detailed prompts for GPT-4, it still uncontrollably generates overly broad or excessively complex guideline steps.
Thus, we employ manual annotation to solve these issues.

Initially, we employ an expert in each domain ({\it{e.g.}}, chef, dancer, etc.) to adjust all guideline steps, aiming to achieve consistent granularity across all of them.
Then, we require the annotators to refine the specific steps generated by GPT-3.5-turbo and annotate the timestamps of steps by watching videos.
It is essential that the refined specific steps contain explicit descriptions of the procedures.
Furthermore, each specific step is also required to be annotated with its corresponding guideline step.

 \begin{figure}[t]
    \centering
    \includegraphics[width=\linewidth]{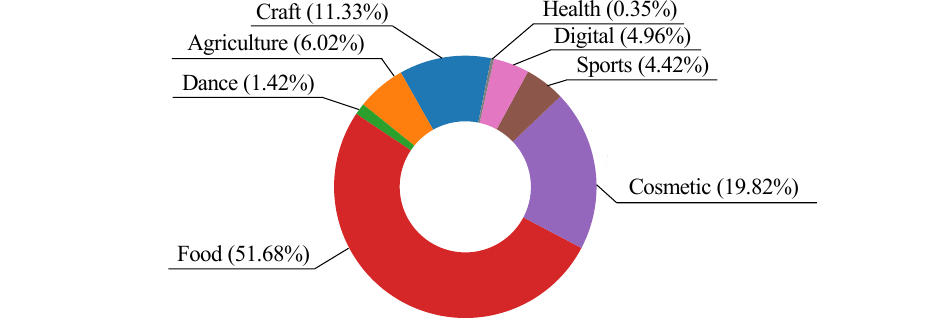}
    \caption{Task category distribution of \textsc{Guide}. There are a wide variety of categories for our videos. The most frequent categories are `Food', `Cosmetic', and `Craft'.}
    \label{fig:static_pie}
\end{figure}

\subsection{Dataset Analysis}
\paragraph{Task Category Distribution}
\textsc{Guide} dataset consists of 560 tasks from 8 common domains in daily life. As shown in Figure~\ref{fig:static_pie}. The top three most frequent domains are `Food', `Cosmetic' and `Craft'.

\paragraph{Dataset Statistics}
We collect a total of 3.5K instructional videos containing 560 different common tasks in daily life. 
The average video duration is 103 seconds, totalling 101 hours. 
Each task contains an average of 6.2 task-related videos and a predefined guideline shared across all task-related videos, resulting in 560 guidelines. 
On average, each guideline consists of 3.7 guideline steps, yielding a total of 2.1K guideline steps with an average length of 2.9 words per guideline step.
Videos are split into multiple segments based on instructional steps, with an average of 4.3 specific steps per video, totalling 15K specific steps. 
Each specific step is annotated with a start-end timestamp and a step caption, with an average length of 6.5 words per caption.
 
\paragraph{Comparisons to Other Datasets}
Table~\ref{table:dataset comparison} compares our \textsc{Guide} dataset to other instructional video datasets.
\textsc{Guide} contains numerous open domain instructional tasks videos, each with annotated specific step captions written by annotators.
While HIREST~\cite{hirest} also provides manually written step captions for open-domain videos, its steps are trivial and unsystematic.
In contrast, each instructional task in \textsc{Guide} is annotated with a guideline, representing a common pattern shared by all task-related videos.
On this basis, we annotate systematic specific steps to provide a clear tutorial.
Moreover, among all datasets containing handwritten step caption, \textsc{Guide} has the largest scale.

\subsection{Task Definition}
\paragraph{Step Captioning}
The step captioning task evaluates the models' capabilities to understand the procedural temporal knowledge of the instructional video.
In this task, models have to generate a set of instructional step captions.
\paragraph{Guideline Summarization}
The guideline summarization task evaluates the models' capabilities to analyze correlations across videos.
In this task, models have to mine the common pattern in task-related videos and summarize a guideline from them.
\paragraph{Guideline-Guided Captioning}
To explore the impact of guidelines on step captioning, we propose the guideline-guided captioning task.
In this task, models have to generate specific step captions under the guide of guideline.

%% file: Tables/table_dataset_comparison.tex
\begin{table*}[h]
\footnotesize
\centering
\resizebox{\linewidth}{!}{
\begin{tabular}{cccccccc}
\toprule
\textbf{Dataset} & \textbf{Duration} & \textbf{Step caption} & \textbf{Guideline-Guided} & \textbf{\#Videos / \# Steps} & \makecell[c]{\textbf{\# Words}\\\textbf{per Caption}} & \makecell[c]{\textbf{\# Steps}\\\textbf{per Video}} & \makecell[c]{\textbf{\# Guideline Steps}\\\textbf{per Task}} \\
\midrule
 COIN~\shortcite{coin}&477h & Predefined& \ding{56}& 11.8K / 46K&4.8 &3.9&- \\
 CrossTask~\shortcite{crosstask}&376h& Predefined& \ding{56}& 4.7K / 21K&2.4 &7.4&- \\
 YouCook2~\shortcite{youcook2}&176h& Manually written & \ding{56}& 2K / 14K&8.8 &7.7&- \\
 HIREST~\shortcite{hirest}&476h& Manually written&\ding{56}& 1.1K / 8.6K&4.4 &7.6&- \\
\midrule
 \textsc{Guide} (Ours)&101h & Manually written (SP) + Predefined (GL)& \ding{52}& 3.5K / 15K & 6.5 &4.3&3.7 \\
\bottomrule
\end{tabular}}
\caption{Comparison of \textsc{Guide} and other datasets with step annotations. Our dataset annotates the common pattern (Guideline Steps) across task-related videos. Moreover, \textsc{Guide} is the largest manually written caption dataset. `SP' Specific Step and `GL' denotes Guideline Step.}
\label{table:dataset comparison}
\end{table*}

%% file: Sections/4_experiments.tex
\section{Experiments}
\input{Tables/table_main_result}
\subsection{Baselines}
\paragraph{Video Foundation Models}
We evaluate three video foundation models on \textsc{Guide}: \textbf{VideoChat}~\cite{videochat}, \textbf{Video-LLaMA}~\cite{videollama} and \textbf{mPLUG-Owl}~\cite{mplug}.
 VideoChat is instantiated using BLIP-2~\cite{blip2} and Vicuna-7B~\cite{vicuna}, and combines pre-trained ViT-G~\cite{vitl} and GMHRA~\cite{gmhra}. 
 Video-LLaMA comprises a pre-trained ViT-G, an audio encoder, Imagebind~\cite{imagebind}, and Vicuna-7B.
 mPLUG-Owl consists of a pre-trained ViT-L~\cite{vitl}, a visual abstractor, and LLaMA-7B~\cite{llama}.

\paragraph{Language Foundation Models}
We evaluate four language foundation models on \textsc{Guide}: \textbf{GPT-3.5-turbo}~\cite{chatgpt}, \textbf{GPT-4}~\cite{gpt4}, \textbf{Flan-T5-XXL}~\cite{flant5} and \textbf{Vicuna-13B}~\cite{vicuna}.
GPT-3.5-turbo and GPT-4 are language models with powerful performance proposed by OpenAI.
Viucuna is a decoder-only architecture and Flan-T5 is an encoder-decoder architecture. 

\paragraph{Human Performance}
To evaluate the gap between the foundation models' comprehension and human understanding of videos, we ask three people (they are familiar with the program but not seen ground-truth annotations) to accomplish this.

\subsection{Evaluation Metrics}
Following previous work~\cite{captionfollow1,hirest}, we use METEOR~\cite{meteor}, CIDEr~\cite{cider} and SPICE~\cite{spice} metrics for evaluating models. 
Additionally, we evaluate the models in two modes for step captioning: Entire Video Captioning (EVC): given an entire video, generate a set of step descriptions.
Video Segment Captioning (VSC): given a ground-truth video segment of the step, generate a text description.
More details are in the Appendix~\ref{app:metrics}.
 
\subsection{Implementation Details}
In video segment captioning (VSC), we divide the video into multiple segments based on ground-truth step timestamps. 
We uniformly sample 8 frames for each segment and feed them to models.
In entire video captioning (EVC), we uniformly sample 32 frames for each video and feed them to models.
In guideline summarization, we modify the input format of the video foundation model to enable simultaneous processing of multiple videos.
We uniformly sample 32 frames from each video as input.
More details are in the Appendix~\ref{app:Implementation Details}.
\input{Tables/table_guideline}
\input{Tables/table_cross_video}
\subsection{Main Results}
The main results are demonstrated in Table~\ref{table:main results}.
We will summarize different findings in the following:
\paragraph{Step Captioning}
We observe from the experimental results that video foundation models demonstrate better performances on video segment captioning (VSC) than entire video captioning (EVC).
This indicates that while the models can comprehend a specific step, they struggle to understand the entire instructional procedure.
One possible explanation is that instructional videos are highly procedural, but the models' pre-training data mainly comprises videos that describe individual events. 
Conversely, language foundation models are not competent for VSC.
This is primarily due to the absence of step-descriptive subtitles in step segments, which hinders models from generating step descriptions based on subtitles. 
 
\paragraph{Guideline Summarization}
We observe from the experimental results that video foundation models are markedly trailing in this task. 
Moreover, even the strong GPT-4 demonstrates a substantial performance gap compared to human beings in this task.
This indicates that these foundation models struggle to mine the correlation across multiple instructional videos.

\paragraph{Guideline-Guided Captioning}
By comparing the results of guideline-guided captioning and step captioning, we observe that both video and language foundation models perform much better with the guide of guidelines.
This demonstrates the helpfulness of the guideline in generating specific steps.
 
\subsection{Analysis}
\paragraph{Importance of Accurate Guideline}
As shown in Table~\ref{tab:guideline}, we use the ground-truth and predicted guideline to guide the generation of specific steps respectively.
The results show that there is significant improvement with the ground-truth guideline, indicating that the guideline is helpful to generate specific steps.
Moreover, the results using the predicted guideline show a substantial decrease.
This further emphasizes the importance of accurate guidelines.

\paragraph{Video Correlation Analysis Capability}
Mining a common guideline from multiple task-related videos depends on the model's capability to analyze correlations across multiple videos.
To investigate the source of this ability, we fine-tune VideoChat (VideoChat has the best fine-tuning performance compared to Video-LLaMA and mPLUG-Owl) under three conditions: (1) single-video: fine-tuning model with single videos along with specific steps ($\text{VideoChat}_{\text{S}}$), (2) multiple-video: fine-tuning model with multiple task-relevant videos along with guideline steps ($\text{VideoChat}_{\text{M}}$), (3) single-video + multiple-video: fine-tuning model under multiple-video setting based on $\text{VideoChat}_{\text{S}}$ ($\text{VideoChat}_{\text{S+M}}$).

As shown in Table~\ref{tab:cross video}, we observe that the performance of  $\text{VideoChat}_{\text{S}}$ and $\text{VideoChat}_{\text{M}}$ 
are superior to 5-shot VideoChat, and the performance of $\text{VideoChat}_{\text{M}}$ is better than $\text{VideoChat}_{\text{S}}$.
However, $\text{VideoChat}_{\text{S+M}}$ shows significant improvement compared to $\text{VideoChat}_{\text{M}}$.
This indicates that the models' ability to analyze correlations across task-related videos is contingent upon their ability to comprehend single-video.
Moreover, models only with single-video comprehension capabilities are incapable of multiple-video comprehension.

 \begin{figure*}[t]
    \centering
    \includegraphics[width=\linewidth]{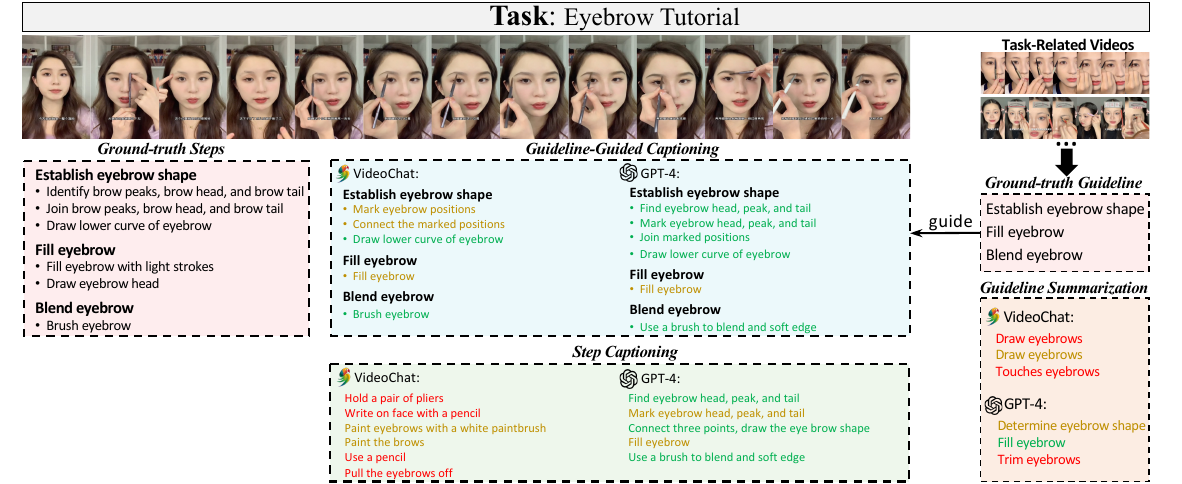}
    \caption{Comparison of foundation models and ground-truth annotation for step captioning, guideline summarization and guideline-guided captioning. Green, yellow, and red text denote `correct', `partially correct', and `wrong' respectively.}
    \label{fig:case}
\end{figure*}

 \begin{figure}[t]
    \centering
    \includegraphics{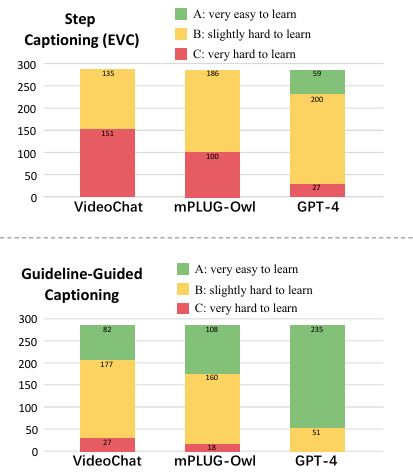}
    \caption{Human evaluation of foundation models on 53 instructional tasks (286 videos).}
    \label{fig:human evl}
\end{figure}

\paragraph{Bottleneck of Video Foundation Models}
To investigate the limitations of video foundation models on instructional video comprehension, we conduct experiments using the mPLUG-Owl on step captioning (EVC). 
We explore three different settings: (1) giving mPLUG-Owl both video and audio (by using video subtitles to simulate audio), (2) giving mPLUG-Owl only video, and (3) giving mPLUG-Owl only audio.
 
The experimental results are shown in Table~\ref{tab:video audio}. 
Surprisingly, mPLUG-Owl shows a significant improvement given only audio compared to when given only video.
Additionally, we observe a substantial drop in the model's performance when adding video after only providing audio. 
We hypothesize that much irrelevant information is mixed in during the visual feature extraction process, which hinders the model's understanding of instructional videos.
This indicates that more specialized visual encoders and visual-language bridges are needed to represent temporal procedures better.

\input{Tables/table_video_audio}
 
\paragraph{Human Evaluation of Foundation Models}
To better evaluate the applicability of the \textsc{Guide} in real-world scenarios, we follow the distribution of the dataset and randomly select 53 instructional tasks (286 videos) for huaman evaluation.
We compare the results of VideoChat, mPLUG-owl, and GPT-4 on step captioning (EVC) and guideline-guided captioning.
The human evaluators are required to rate the output based on whether the steps are clear and easy to learn.
We implemented a three-level rating system to categorize the quality of outputs.
A means `steps are very easy to learn', B means `steps are slightly hard to learn', and C means `steps are very hard to learn'.
As shown in Figure~\ref{fig:human evl}, guideline-guided captioning has better results compared to step captioning, indicating that the guideline helps models generate clearer and easier-to-learn instructional steps.

\subsection{Case Study}

In Figure~\ref{fig:case}, we list an example generated by foundation models and ground-truth annotation for step captioning (EVC), guideline summarization and guideline-guided captioning task on videos associated with the `Eyebrow Tutorial’.
In the step captioning (EVC), the specific steps generated by VideoChat are very inaccurate.
The powerful GPT-4 also has some missing and redundant steps.
In the guideline summarization, both GPT-4 and VideoChat are unsuccessful in summarizing the accurate guideline.
In the guideline-guided captioning, the VideoChat results show a significant improvement in clarity and accuracy, and the specific steps generated by GPT-4 are essentially entirely correct.

%% file: Tables/table_main_result.tex
\begin{table*}[h]
	\centering
	\label{tab:mainexperiments}
	\resizebox{\linewidth}{!}{
		\begin{tabular}{lcccccccccccc}
			\toprule
                \multirow{2}{*}{Methods} & & \multicolumn{3}{c}{Step Captioning (EVC / VSC)} & & \multicolumn{3}{c}{Guideline Summarization} & & \multicolumn{3}{c}{Guideline-Guided Captioning} \\ \cmidrule{3-5} \cmidrule{7-9} \cmidrule{11-13}
			& & METEOR & CIDEr & SPICE & & METEOR & CIDEr & SPICE & & METEOR & CIDEr & SPICE\\
			\midrule
                Human Performance &&22.5 / 26.0 &65.6 / 78.5 &24.1 / 38.6 &&13.6 &56.6  &14.0 &&31.8 &73.4  &36.9  \\
                \midrule
                \multicolumn{13}{c}{(a) \textit{Video Foundation Models}} \\ \midrule
			$\text{VideoChat}_{\text{5-shot}}$ &&\underline{6.8} / 4.2 &1.8 / \underline{4.7} &3.6 / \underline{3.7} &&3.8 &0.1 &2.1 &&8.8 &4.3 &5.6  \\
      		$\text{Video-LLaMA}_{\text{5-shot}}$ &&\underline{4.1} / 2.3 &1.1 / \underline{1.3}  &\underline{1.7} / 0.9 &&2.8 &0.2 &0.7 &&8.2 &2.3 &2.6 \\
      		$\text{mPLUG-Owl}_{\text{5-shot}}$ &&\underline{7.9} / \textbf{5.8}  &6.4 / \textbf{\underline{9.6}} &5.9 / \textbf{\underline{6.8}} &&2.2 &2.4  &1.4 &&9.1 &8.6  &7.5\\
                \midrule
                \multicolumn{13}{c}{(b) \textit{Language Foundation Models}} \\ \midrule
                $\text{Flan-T5}_{\text{5-shot}}$ && 4.7 / - &  1.9 / -  & 5.2 / -  &&3.3 &3.6 &1.4  &&12.9 &7.4 &11.6\\
                $\text{Vicuna}_{\text{5-shot}}$ && 9.5 / - &  4.5 / -  & 7.4 / -   &&6.3 &5.0 &4.9  &&11.5 &7.8 &9.3\\
      		$\text{GPT-3.5-turbo}_{\text{5-shot}}$ &&  14.9 / - &  11.2 / - &  12.4 / -  &&9.4 &13.3 &9.3 &&17.2 &13.1 &13.3\\
                $\text{GPT-4}_{\text{zreo-shot}}$ && 16.7 / - &  5.9 / -  & 12.1 / -  &&9.9 &19.5 &6.8 &&22.8 &16.0 &18.7\\
      		$\text{GPT-4}_{\text{5-shot}}$ && \textbf{16.8} / -  & \textbf{13.1} / -  & \textbf{13.2} / -  &&\textbf{10.4} &\textbf{24.5} &\textbf{9.6}  &&\textbf{23.5} &\textbf{18.8} &\textbf{21.9} \\    
                \bottomrule
		\end{tabular}
	}
	\caption{The results on three sub-tasks. We report the average results of three runs. `EVC' denotes entire video captioning and `VSC' denotes video segment captioning. Best results in each task are highlighted by \textbf{bold}. Better results between EVC and VSC are highlighted by \underline{underline}.}
	\label{table:main results}
\end{table*}

%% file: Tables/table_guideline.tex
\begin{table}[h]
    \centering
    \resizebox{\linewidth}{!}{
    \begin{tabular}{lccc}
        \toprule
        \multirow{2}{*}{Method} & \multicolumn{3}{c}{Guideline-Guided Captioning} \\ \cmidrule{2-4} 
        &METEOR  & CIDEr & SPICE \\
        \midrule
        $\text{mPLUG-Owl}$  &7.9 &6.4 &5.9 \\
        $\text{\it \quad- Pred-guideline}$   &4.7 &2.6 &3.1 \\
        $\text{\it \quad- GT-guideline}$   &\textbf{9.1} &\textbf{8.6} &\textbf{7.5} \\
        \bottomrule
    \end{tabular}}
    \caption{The results of the mPLUG-Owl on guideline-guided captioning task with predicted and ground-truth guideline inputs. The first line is the result of the model without guideline. Best results are highlighted by \textbf{bold}.}
    \label{tab:guideline}
\end{table}

%% file: Tables/table_cross_video.tex
\begin{table}[h]
    \centering
    \scalebox{1.0}{
    \begin{tabular}{lccc}
        \toprule
        \multirow{2}{*}{Method} & \multicolumn{3}{c}{Guideline Summarization} \\ \cmidrule{2-4} 
        &METEOR  & CIDEr & SPICE\\
        \midrule
        $\text{VideoChat}_{\text{S+M}}$  &\textbf{7.3} &\textbf{14.9}  &\textbf{5.8}       \\
        $\text{VideoChat}_{\text{S}}$  &\underline{7.2} &\underline{10.8}  &\underline{3.4}  \\ 
        $\text{VideoChat}_{\text{M}}$  &3.4 &2.5  &2.9  \\ 
        $\text{VideoChat}_{\text{5-shot}}$  &3.8 &0.1  &2.1       \\
        \bottomrule
    \end{tabular}
    }
    \caption{The results for VideoChat on guideline summarization task under different fine-tuning settings. Best and second results are highlighted by \textbf{bold} and \underline{underline}.}
    \label{tab:cross video}
\end{table}

%% file: Tables/table_video_audio.tex
\begin{table}
    \centering
    \resizebox{\linewidth}{!}{
    \begin{tabular}{lccc}
        \toprule
        \multirow{2}{*}{Method} & \multicolumn{3}{c}{Step Captioning (EVC)} \\ \cmidrule{2-4} 
        &METEOR  & CIDEr & SPICE \\
        \midrule
        $\text{mPLUG-Owl}_{\text{Video+Audio}}$  &\underline{5.5} &\underline{3.2} &\underline{3.1} \\
        $\text{mPLUG-Owl}_{\text{Video}}$   &3.6 &2.9 &2.1 \\
        $\text{mPLUG-Owl}_{\text{Audio}}$   &\textbf{8.2} &\textbf{6.1} &\textbf{6.8} \\
        \bottomrule
    \end{tabular}}
    \caption{The results of the $\text{mPLUG-Owl}_{\text{zero-shot}}$ on step captioning (EVC) task with different modal information inputs. Best and second results are highlighted by \textbf{bold} and \underline{underline}.}
    \label{tab:video audio}
\end{table}

%% file: Sections/5_conclusion.tex
\section{Conclusion}
In this paper, we introduce a guideline-guided dataset (\textsc{Guide}) for instructional video comprehension and propose three sub-tasks based on the \textsc{Guide}. 
We evaluate various foundation models on our dataset. 
Experimental results show that the accurate guideline is beneficial for generating clear, easier-to-learn and systemic specific steps.
With different training settings, we found that the key to extracting an accurate guideline from multiple videos is single-video understanding ability, indicating more pre-training and fine-tuning data are necessary.
Moreover, we observed in the ablation study that the bottleneck for video foundation models is the visual modality. 
We believe more specialized visual encoders and visual-language bridges are needed to to represent instructional procedures better.
Finally, we perform a human evaluation demonstrating our dataset's promising applications in real-world scenarios.

%% file: Sections/6_ethics.tex
\section*{Ethics Concerns}
The videos are sourced from an open-source video platform that we cooperate with.
Users have agreed to transfer the copyright to the platform when uploading videos, which does not involve privacy issues.
Moreover, videos have been subject to strict ethical review (non-ethical content and sensitive information) by the platform.

%% file: Sections/7_appendix.tex
\section{Appendix}
\subsection{Dataset Construction Pipeline}\label{app:Dataset Construction Pipeline}
This dataset construction process took about 400 hours, with 16 professional annotators and 3 quality inspectors.
Before the start of each annotation stage, we provide targeted training to the annotators.
To ensure they understand our annotation goals, each stage is divided into two processes: Trial Annotation and Formal Annotation. 
Formal Annotation begins by assuming that the results of Trial Annotation meet our expectations.
At the end of the Annotation, inspectors check the annotation results to determine the annotators' salary.

\subsubsection{Video Collection.}\label{app:Video Collection}
In this stage, we require that the collected videos have clear operational procedures and a detailed voice explanation accompany each procedure.
To improve the efficiency, we first perform a preliminary selection of the video based on the length of the video transcription. 
We remove videos with transcribed text lengths of less than 80 characters (possibly excluding voice explanation or containing insufficiently detailed voice explanation).
Then, we require the annotators to oversee the videos and collect videos that have clear procedures and detailed voice explanations.
Moreover, the volume of the background music is required to be lower than the volume of the voice explanation.
The total number of videos we collected in the video database is 13.8K.
However, due to the database’s poor quality, some irrelevant videos may have been included in the task queries.
These videos were filtered out during the automatic annotation stage, resulting in a total of 3.5K videos retained.

\subsubsection{Automatic Annotation.}\label{app:Automatic Annotation}
\begin{figure*}[t]
    \centering
    \includegraphics[width=\linewidth]{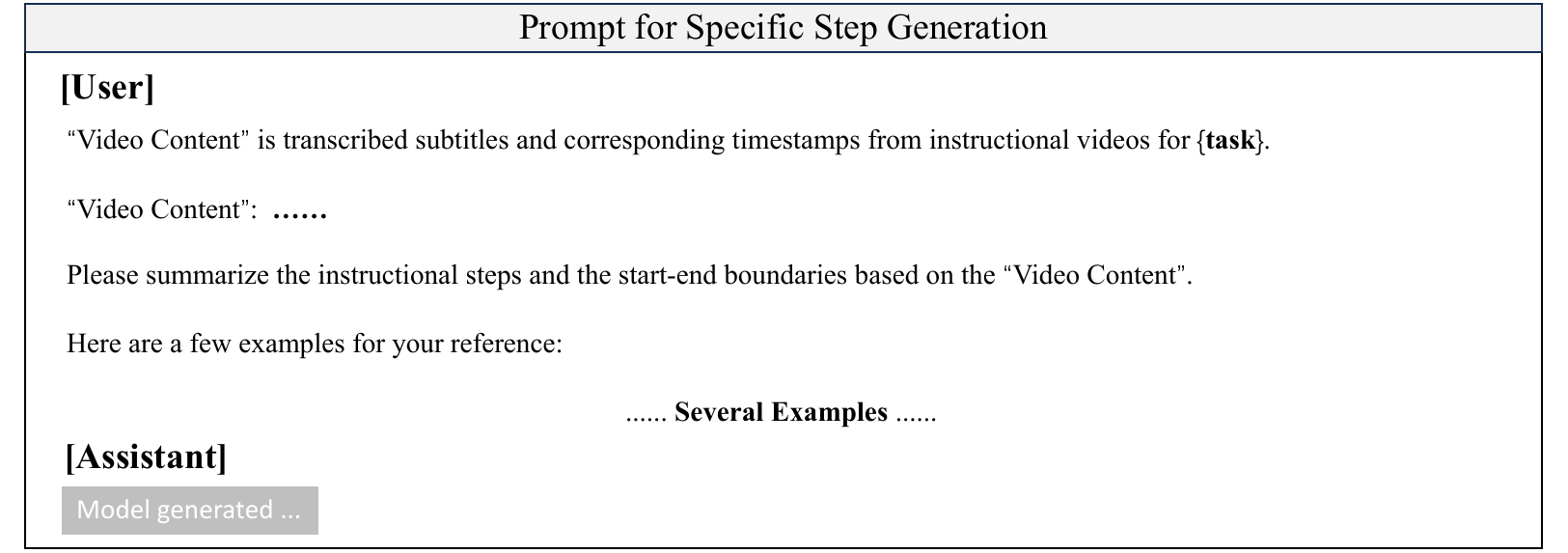}
    \caption{The prompt for generating specific steps.}
    \label{fig:specific}
\end{figure*}

\begin{figure*}[t]
    \centering
    \includegraphics[width=\linewidth]{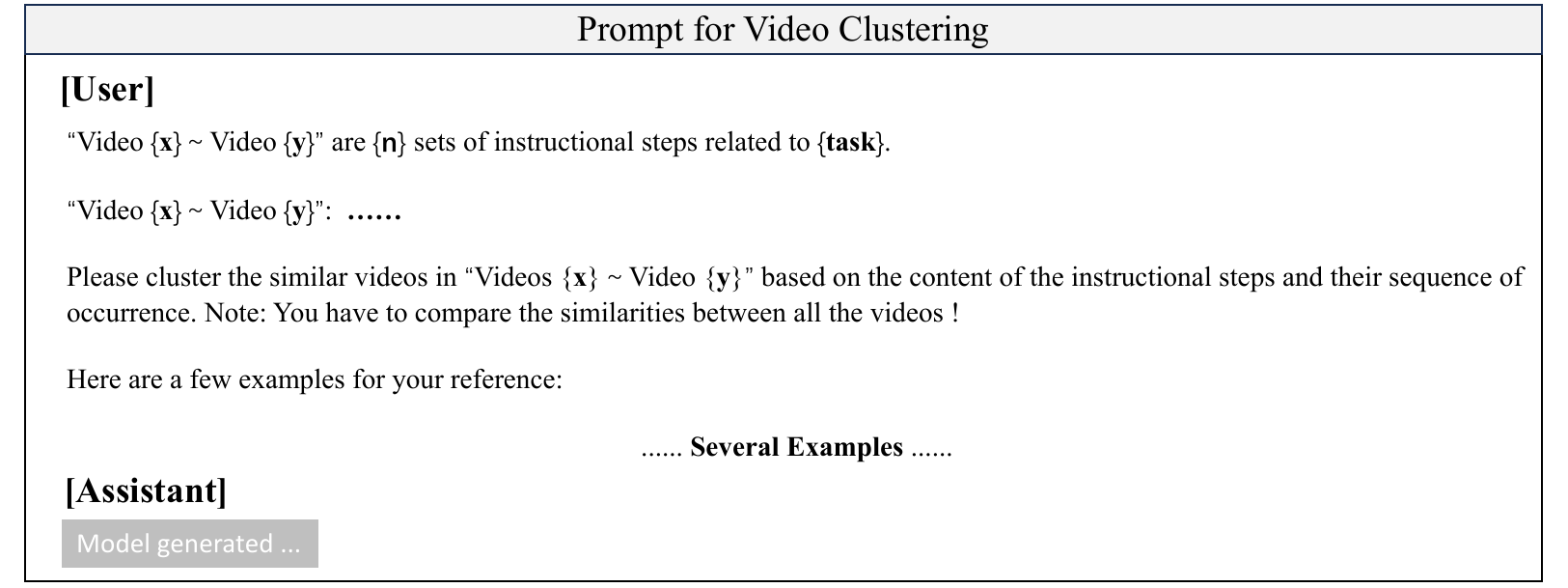}
    \caption{The prompt for clustering videos.}
    \label{fig:cluster}
\end{figure*}

\begin{figure*}[t]
    \centering
    \includegraphics[width=\linewidth]{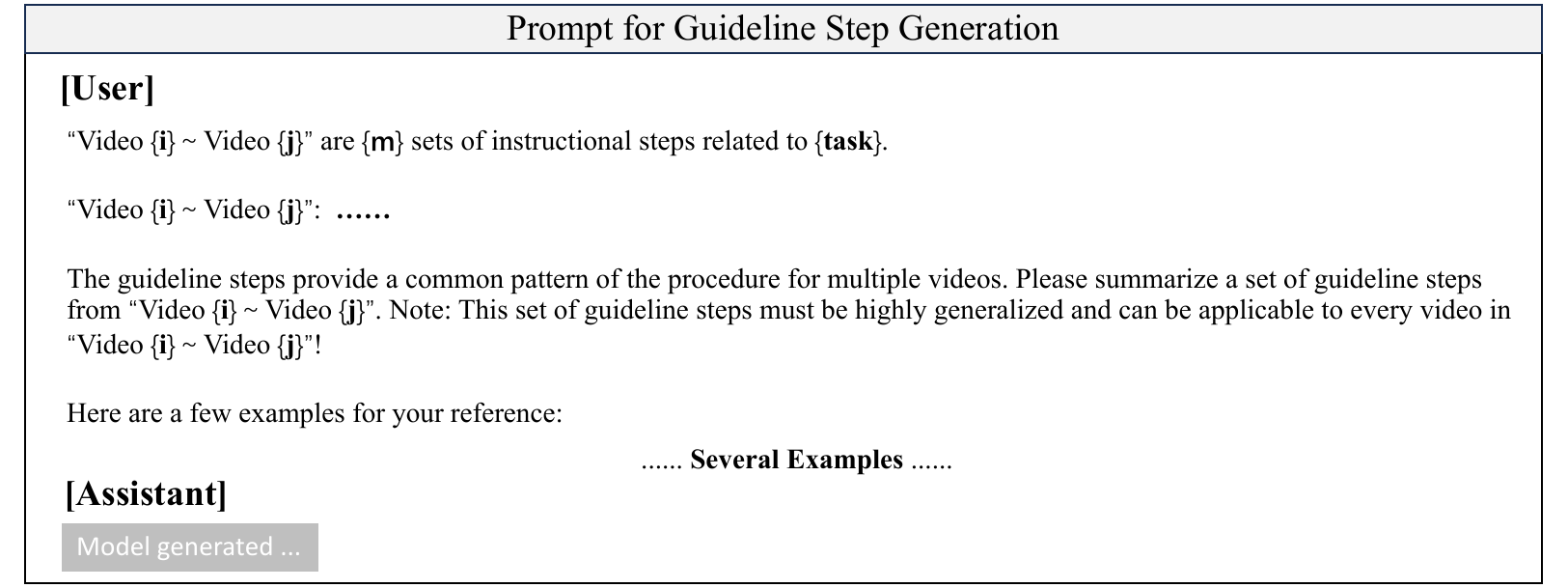}
    \caption{The prompt for generating guideline steps.}
    \label{fig:outline}
\end{figure*}

In this stage, we carefully craft three prompts. These prompts are employed to generate specific steps (shown in Figure~\ref{fig:specific}), cluster videos (shown in Figure~\ref{fig:cluster}), and generate guideline steps for task-related videos (shown in Figure~\ref{fig:outline}).

\subsubsection{Quality of Automatic Annotations.}
We randomly sample 20 queries (112 videos) in the dataset. We list the automatic annotations and the manually refined annotations.
We ask 3 volunteers who had not seen ground-truth annotations to vote for the more accurate one, i.e., manually refined annotations.
In the end, 99\% (111 videos) of the videos are judged correctly, proving the lack of reliability of automatic annotations.

\subsubsection{Efficiency Improvements with Automatic Annotation.}
We initially relied on manual annotation, which proved to be highly inefficient.
Annotating 50 queries (about 320 videos) took annotators 80 hours, making them 2 times less efficient than the current solution.

\subsubsection{Manual Annotation.}
In the specific step generation stage, GPT-3.5-turbo clusters videos based on specific steps, aiming to group multiple task-related videos that can share the same set of guideline steps.
 However, despite trying various prompts, certain videos still do not suit this set of guideline steps. 
 Therefore, we require the annotators to watch clustered task-related videos and filter out those that are not suitable.
 In refining the specific steps, we provide annotators with timestamps generated by GPT-3.5-turbo. 
 This reference assists annotators in quickly locating the specific steps in the videos, thus enhancing the efficiency of the annotation process.

\subsection{LLM-based Evaluation Strategies.}\label{app:metrics}
We use GPT-4 to evaluate the performance of the VideoChat in the Guideline Summarization task.
The evaluation samples and rules are consistent with the human evaluation.
In the evaluation results, 17 samples are inconsistent with human evaluation results (17/53).
Therefore, we consider the GPT evaluation unreliable.

\subsection{Implementation Details}\label{app:Implementation Details}
\begin{figure}[t]
    \centering
    \includegraphics[width=\linewidth]{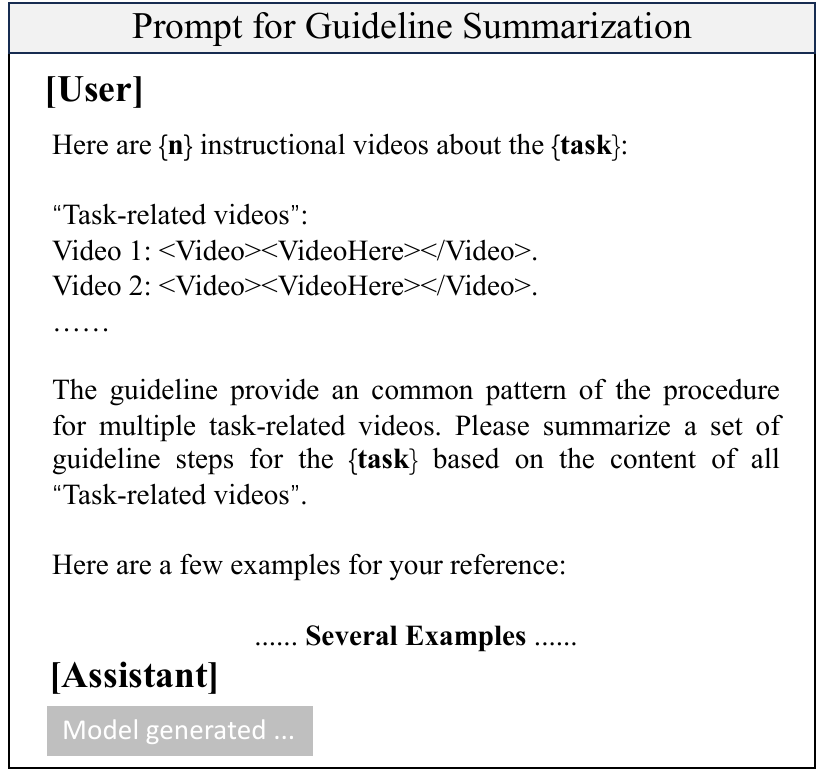}
    \caption{The prompt for guideline summarization task.}
    \label{fig:summ}
\end{figure}

We mainly follow their original settings to evaluate video foundation models.
 The VideoChat is fine-tuned on a single Tesla A100 80G GPU for 5 epochs. 
 The rest of the hyperparameters are the same as the original VideoChat.
 To enable the video foundation models to process multiple videos simultaneously for the guideline summarization task, we design a multiple-video prompt (shown in Figure~\ref{fig:summ}).

